\documentclass[11pt]{article}

\usepackage[preprint]{acl}

\usepackage{times}
\usepackage{latexsym}
\usepackage{booktabs}
\usepackage{amsmath}
\usepackage{enumitem}
\usepackage{float}
\usepackage{nameref}
\usepackage[T1]{fontenc}

\usepackage[utf8]{inputenc}

\usepackage{microtype}
\usepackage{subcaption}

\usepackage{inconsolata}

\usepackage{graphicx}

\title{Can Frontier LLMs Match Natively Multimodal Embeddings? A Comparison on Hard-Negative Text-to-Image Retrieval}

\author{Archan Dutta \\
  Westcliff University \\
  \texttt{a.dutta.171@westcliff.edu} \\\And
  Vyanktesh Kanungo \\
  \texttt{vyankteshkanungo@gmail.com } \\}

\begin{document}
\maketitle
\begin{abstract}
Multimodal retrieval and classification across different types of media, spanning text, images, video and audio, has traditionally relied on dual-encoder models that align visual and textual representations through contrastive learning. The March 2026 release of Gemini Embedding 2, Google's first natively multimodal embedding model to map text, images, video, audio, and documents into a single shared space, raises competition among multimodal retrieval systems. Simultaneously, frontier Large language models (LLMs) have also demonstrated strong visual understanding, raising the question of whether they can serve as effective zero-shot rankers. Our study provides the first direct comparison of native multimodal embeddings against LLM-based visual ranking on Flickr30k. We observe that GPT-4.1 and Claude Sonnet 4.6 perform on par with Gemini Embedding 2.  Additionally, once embeddings are precomputed, multimodal embeddings are better suited for low-latency applications.
\end{abstract}

\section{Introduction}

Text-image retrieval, ranking a gallery of images by relevance to a natural language query, is a long-standing task with applications in visual search, content recommendation, and retrieval-augmented generation. There have been two recent developments in this space. First, LLMs with strong vision capabilities, such as GPT-4.1 \citep{openai2025gpt41} and Claude \citep{anthropic2024claude}, can now receive images and text jointly, enabling a fundamentally different retrieval paradigm: presenting all candidate images alongside the query in one prompt and ranking them directly. Second, natively multimodal embedding models have recently become publicly available: Gemini Embedding 2 (GE2) \citep{google2026ge2} and Amazon Nova 2 (Nova 2) \citep{amazon2025novamme}, released in March, 2026 and October 2025, respectively, both project multiple modalities through a single shared backbone into a unified embedding space, unlike dual-encoder models that align independently encoded representations.
The comparison of native multimodal embedding (GE2, Nova 2) with direct LLM visual ranking (GPT-4.1, Claude Sonnet 4.6) is timely and currently uncharacterized in the literature, specifically on (1) \textbf{Retrieval Accuracy} and (2) \textbf{Ranking Time}
Our contributions are as follows:
\begin{itemize}\itemsep0pt
\item The first comparison of natively multimodal embedding models against frontier LLMs as zero-shot direct visual rankers on Flickr30k \citep{young2014flickr30k}, using semantically challenging hard-negative candidate sets without task-specific fine-tuning or intermediate caption generation.
\item An empirical finding that GE2, GPT-4.1, and Claude Sonnet 4.6 achieve statistically indistinguishable retrieval accuracy, and embedding models rank 1000 queries in under two seconds (with pre-computation) compared to over three hours for LLMs.
\end{itemize}

\section{Related Work}
\textbf{Dual-encoder cross-modal retrieval.}
CLIP \citep{radford2021clip} established the paradigm of learning aligned image and text encoders via large-scale contrastive pretraining, achieving strong zero-shot transfer on image-text retrieval benchmarks. ALIGN \citep{jia2021align} demonstrated that data scale can compensate for annotation quality. More recent work, including BLIP-2 \citep{li2023blip2} and EVA-CLIP \citep{fang2023evaclip}, has extended these ideas with larger architectures and richer pretraining objectives, achieving state-of-the-art performance on COCO \citep{lin2014coco} and Flickr30k.
\vspace{-1pt}
\textbf{LLMs for retrieval and re-ranking.}
\citet{sun2023chatgpt} found that ChatGPT achieves competitive performance as a zero-shot document ranker on text-only benchmarks. \citet{ma2023zero} showed that prompting LLMs to produce listwise rankings outperforms pointwise scoring in text retrieval. In the multimodal setting, BLIP-2 \citep{li2023blip2} and InstructBLIP \citep{dai2023instructblip} use LLMs to generate textual image descriptions, which are then matched against queries by a text retrieval system. Our work departs from this indirect approach by prompting LLMs to rank all candidate images together. MM-Embed \citep{jiang2024mmembed} fine-tunes a multimodal LLM-based retriever but focuses on learning retrieval representations rather than evaluating frontier models as zero-shot rankers. RagVL \citep{zhu2024mllm} demonstrates that multimodal LLMs are strong rerankers in retrieval-augmented generation settings, though in a document QA context rather than image retrieval.

\noindent\textbf{Natively multimodal embeddings.}
ImageBind \citep{girdhar2023imagebind} demonstrated joint embedding of six modalities using image-paired data as a binding signal, though its underlying encoders remain modality-specific. VLM2Vec \citep{jiang2024vlm2vec} takes a step further by fine-tuning a vision-language model as a unified embedding backbone, leveraging deep integration of vision and language features within a single transformer architecture rather than late fusion of independently encoded representations.

\noindent\textbf{Hard negative mining.}
VSE++ \citep{faghri2018vse} introduced hard negative mining as a key training-time improvement, showing that selecting the most violating negatives substantially improves rank-based metrics. \citet{robinson2021contrastive} also demonstrate that semantically hard negatives substantially change task difficulty.

\section{Methodology} \label{sec:methodology}

\subsection{Dataset, Sampling and Evaluated Systems}
Flickr30k \citep{young2014flickr30k} provides 31{,}000 images each annotated with five independent captions. Although full-gallery retrieval on Flickr30k is largely saturated at the top of the leaderboard, our hard-negative candidate sets create a non-trivial task. We sample a pool of 5{,}000 images from Flickr30k, each paired with five independent human-written captions. To obtain a single
representative query caption per image while controlling for annotator-style
variance, we apply a \emph{stratified} caption sampling: the 5{,}000 images are partitioned into five equally-sized groups of 1{,}000, each assigned exclusively to one caption index (0 to 4). The ground truth for every query is the image from which its caption was originally drawn.

\begin{figure}[t]
\centering
\includegraphics[width=0.82\columnwidth]{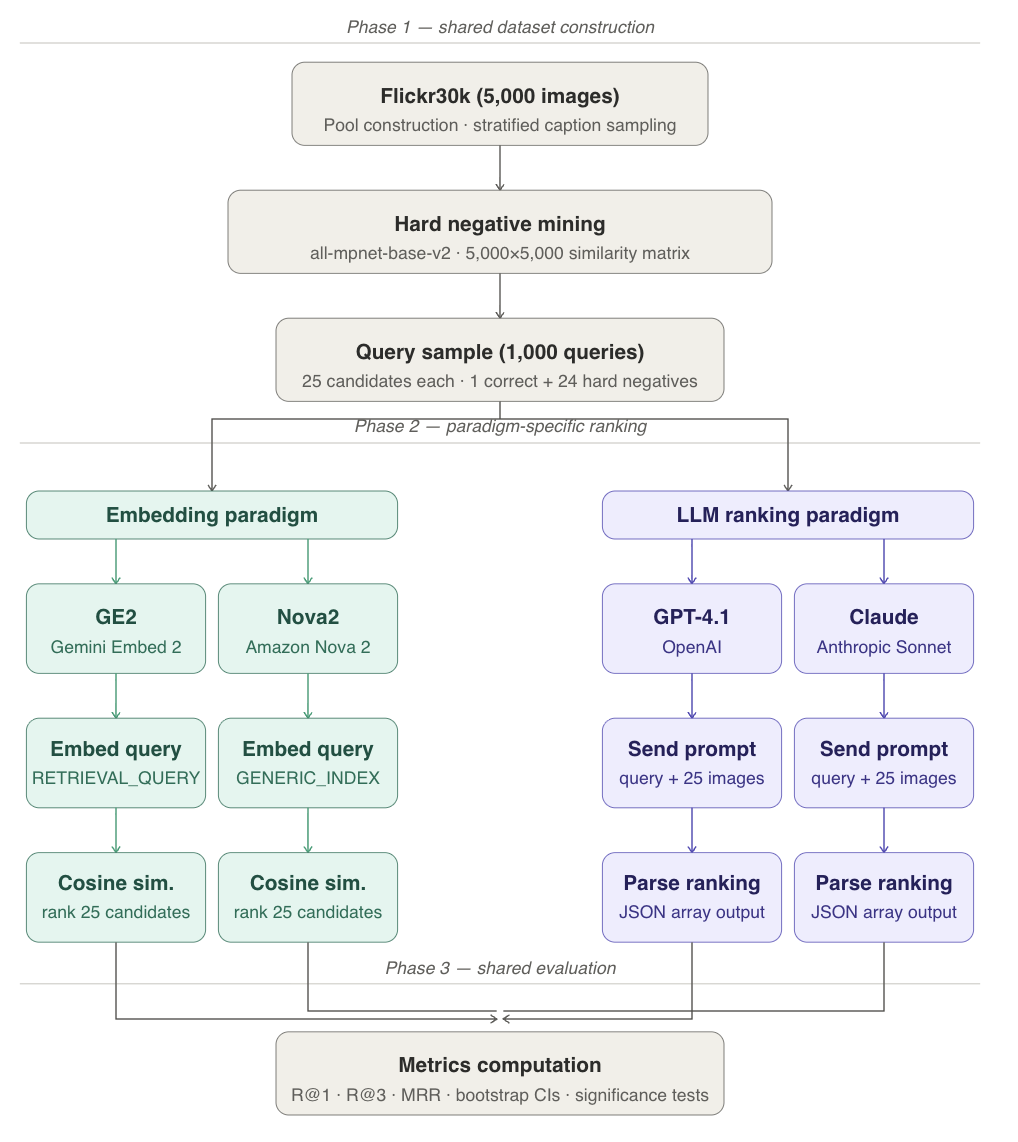}
\caption{Overview of the three-phase evaluation pipeline: hard negative dataset construction, ranking via embedding models and LLMs, and evaluation.}
\label{fig:pipeline}
\end{figure}

\subsection{Hard Negative Candidate Set Construction}
For each of the 5{,}000 pool images, we constructed a candidate set of 25
images: the ground-truth image and the 24 hardest negatives. This candidate set size relatively balances task difficulty with the practical constraint of fitting all images within a single LLM prompt without exceeding context limits.
Hard negatives are selected by embedding all representative captions with
\texttt{sentence-transformers/all-mpnet-base-v2}
\citep{reimers2019sbert}, a general-purpose text encoder that is intentionally selected to be decoupled from all evaluated retrieval systems. We then compute a 5{,}000$\times$5{,}000 pairwise cosine similarity matrix, and retain the 24 most similar captions to the query (excluding the query itself).
The images corresponding to these captions form the hard negative set. From the full pool, we randomly sample 1{,}000 captions as queries. Figure \ref{fig:pipeline} shows the pipeline diagram.
 

\begin{figure}[t]
\centering
\includegraphics[width=0.48\columnwidth]{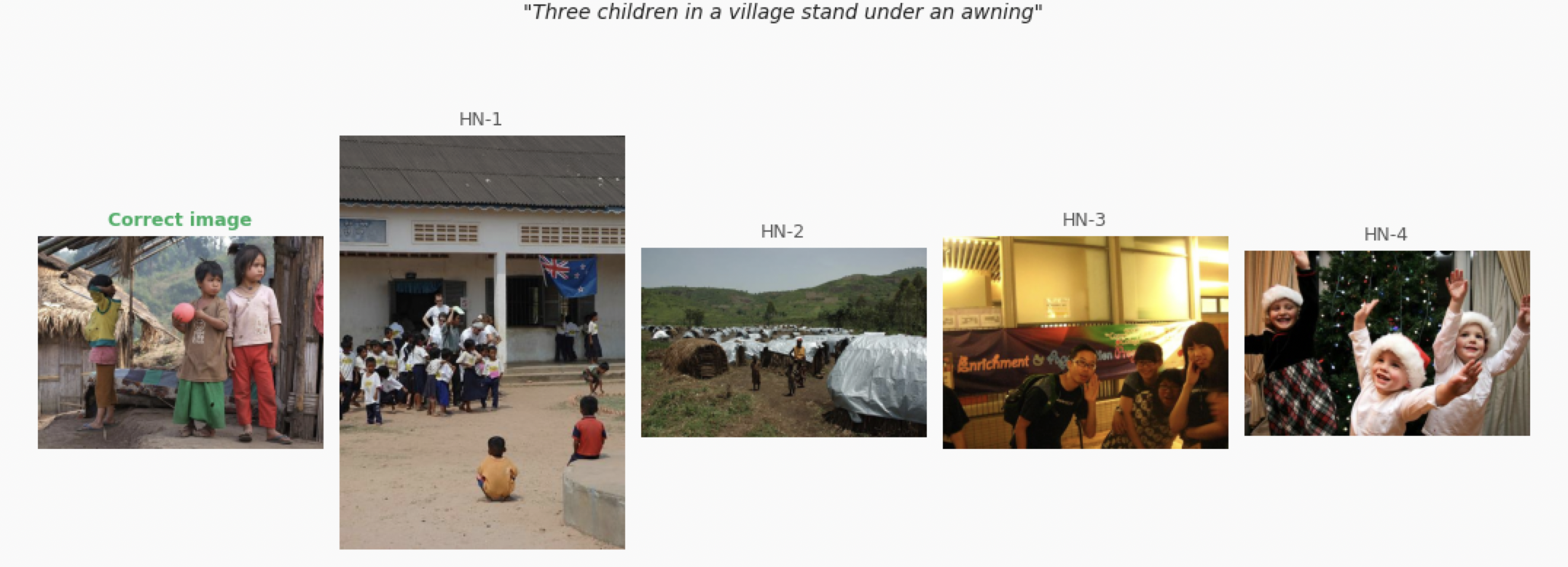}
\hfill
\includegraphics[width=0.48\columnwidth]{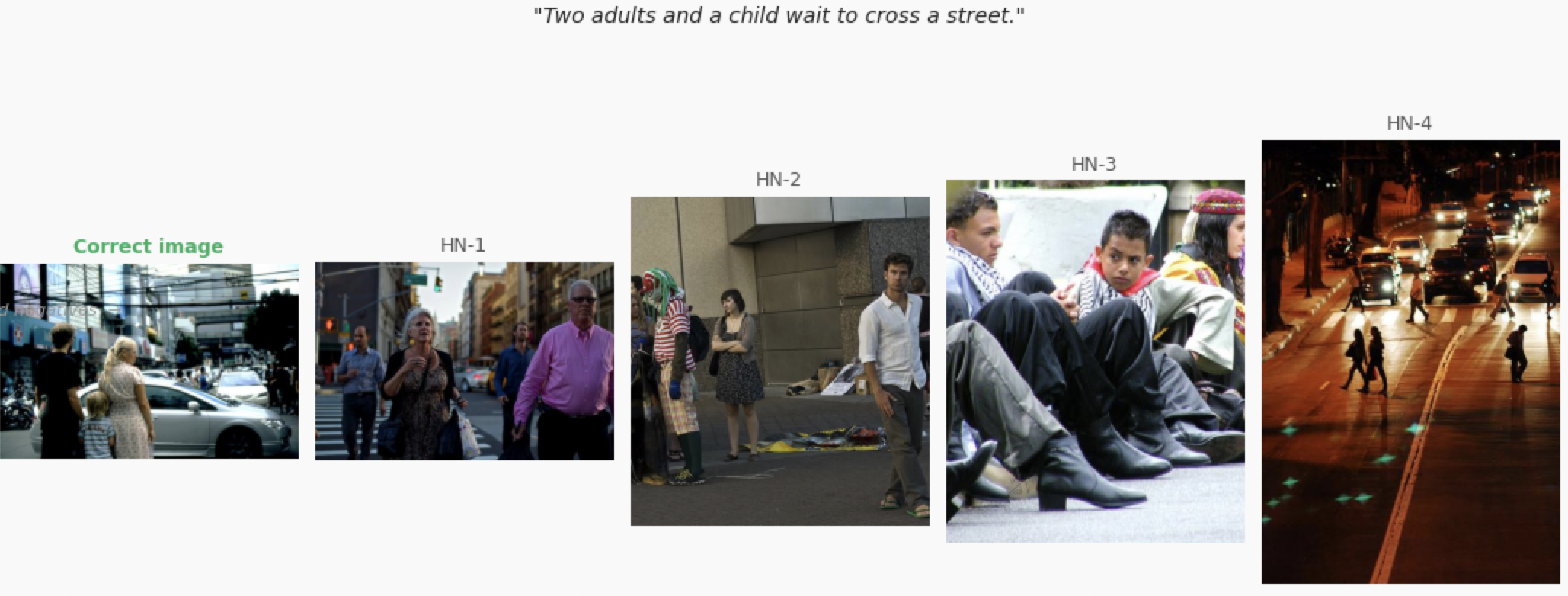}
\caption{Examples of hard negative datasets showing candidate images for a given query.}
\label{fig:hard_negatives}
\end{figure}

We construct hard negatives in \emph{caption} embedding space rather than image embedding space for two reasons. First, grounding distractor difficulty in caption-space similarity increases ambiguity, making images hard to distinguish \emph{given the caption}, which is precisely the retrieval difficulty we wish to evaluate. Second, since human perception of image similarity is naturally expressed through language, using caption similarity to construct candidate sets is a well-motivated design choice (details mentioned in \nameref{sec:limitations}). We note that constructing hard negatives in caption space may favor embedding-based retrieval methods. All evaluated systems (Gemini Embedding 2, Amazon Nova 2 Multimodal Embedding, GPT-4.1, and  Claude Sonnet 4.6) are exposed to identical queries and candidate sets. Details about system configurations are provided in Appendix \ref{appendix:systems_configuration}

\subsection{Evaluation Metrics and Statistical Tests}
We report three metrics over 1{,}000 queries:
\begin{itemize}\itemsep 0pt
  \item \textbf{Recall@1 (R@1)}: Fraction of queries where the ground-truth image is ranked first.
  \item \textbf{Recall@3 (R@3)}: Fraction of queries where the ground-truth image appears in the top three.
  \item \textbf{Mean Reciprocal Rank (MRR)}: Mean of the reciprocal rank of the ground-truth image across all queries.
\end{itemize}
Additionally, we report 95\% bootstrap confidence intervals for all three metrics (Appendix~\ref{appendix:metrics_with_ci}) using 1,000 resamples \citep{efron1993bootstrap}. For pairwise
significance testing of R@1 (a binary outcome), we apply McNemar's test \citep{mcnemar1947note}. For MRR (a continuous outcome), we apply
the Wilcoxon signed-rank test \citep{wilcoxon1945individual}. We use a
significance threshold of $\alpha = 0.05$ with Bonferroni correction for three pairwise comparisons.

 


\section{Results and Discussion}
\label{sec:results-and-discussion}

\begin{figure}[t]
  \centering
  \includegraphics[width=\columnwidth]{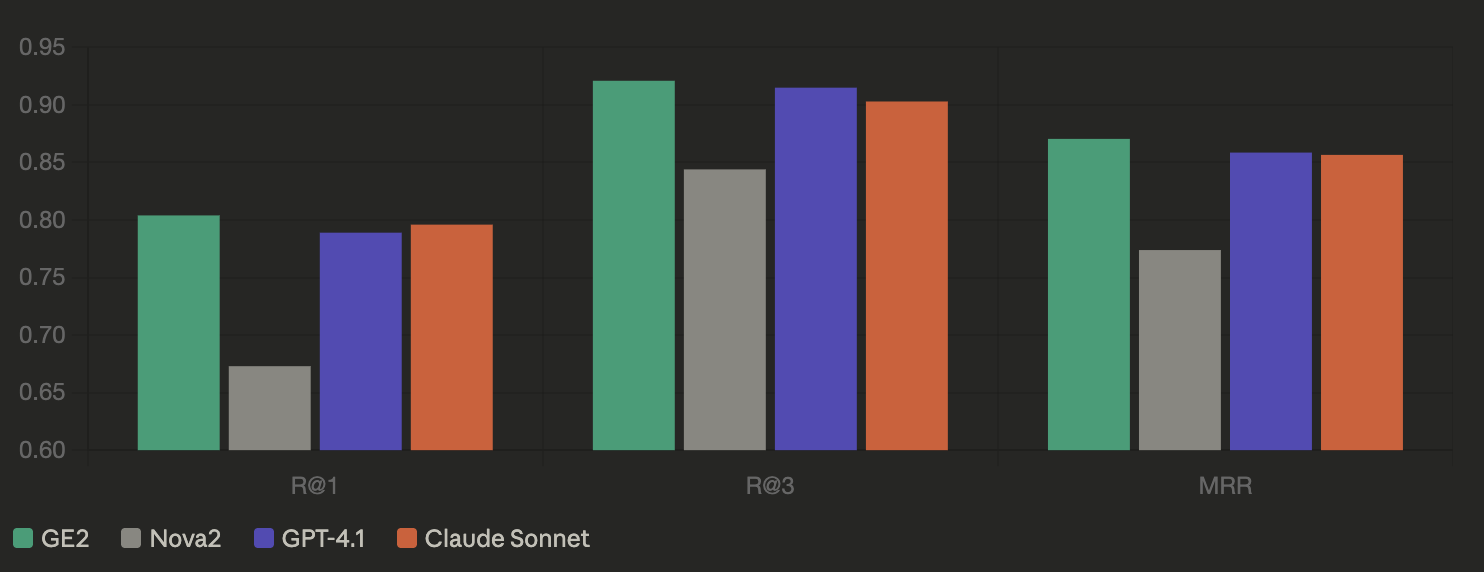}
  \caption{Evaluating the four systems on Recall@1, Recall@3 and MRR.}
  \label{fig:ACL_Multimodal_Primary_Result}
\end{figure}

Figure~\ref{fig:ACL_Multimodal_Primary_Result} visualizes the metric comparison. GE2 achieves the highest score across all three metrics, followed closely by Claude Sonnet 4.6 and GPT-4.1, with Amazon Nova 2 trailing by 13 percentage points. However, pairwise significance testing (Table~\ref{tab:significance}) reveals a clear separation: Gemini Embedding 2, GPT-4.1, and Claude Sonnet 4.6 are statistically indistinguishable from one another on both R@1 (McNemar, all p > 0.25) and MRR (Wilcoxon, all p > 0.09), while the performance of Amazon Nova 2 is significantly different from every other system (p < 0.0001 on both tests). The absence of a statistically significant gap between the embedding paradigm (Gemini Embedding 2) and the LLM reasoning paradigm (GPT-4.1, Claude Sonnet 4.6) is the primary finding of this study (given the dataset and candidate set). Dedicated multimodal embedding and joint visual reasoning achieve competitive retrieval accuracy on hard-negative candidates drawn from a large image pool.


\begin{table*}[t]
\centering
\small

\begin{tabular}{llcccc}
\toprule
\textbf{System A} & \textbf{System B} & \textbf{Recall@1 - McNemar $p$} & \textbf{Sig.} & \textbf{MRR - Wilcoxon $p$} & \textbf{Sig.} \\
\midrule
Gemini Embedding 2 & GPT-4.1       & 0.2699 & No  & 0.1574 & No  \\
Gemini Embedding 2 & Claude Sonnet 4.6 & 0.5569 & No  & 0.0995 & No  \\
GPT-4.1            & Claude Sonnet 4.6 & 0.5854 & No  & 0.9354 & No  \\
\midrule
Gemini Embedding 2 & Amazon Nova 2 & $<$0.0001 & \textbf{Yes} & $<$0.0001 & \textbf{Yes} \\
GPT-4.1            & Amazon Nova 2 & $<$0.0001 & \textbf{Yes} & $<$0.0001 & \textbf{Yes} \\
Claude Sonnet 4.6      & Amazon Nova 2 & $<$0.0001 & \textbf{Yes} & $<$0.0001 & \textbf{Yes} \\
\bottomrule
\end{tabular}
\caption{Pairwise statistical significance tests. McNemar's test is applied to paired binary R@1 outcomes; Wilcoxon signed-rank test is applied to paired MRR scores. The upper block shows non-significant pairs; the lower block shows pairs involving Amazon Nova 2, all significant at $p < 0.0001$.}
\label{tab:significance}

\vspace{8pt}

\setlength{\tabcolsep}{4pt}
\begin{tabular}{lcccc}
\toprule
 & \multicolumn{3}{c}{\textbf{Precomputation Time}} \\
\cmidrule(lr){2-4} 
\textbf{System} & \textbf{Query emb.} & \textbf{Image emb.} & \textbf{Total Time} & \textbf{Ranking-only Time} \\
\midrule
Gemini Embedding 2
  & ${\sim}2{,}000$\,s
  & ${\sim}12{,}500$\,s
  & ${\sim}14{,}500$\,s
  & \textbf{0.66\,s} (with precomp.) \\
Amazon Nova 2
  & ${\sim}1{,}800$\,s
  & ${\sim}10{,}000$\,s
  & ${\sim}11{,}800$\,s
  & \textbf{1.16\,s} (with precomp.) \\
\midrule
GPT-4.1
  & \multicolumn{3}{c}{N/A (end-to-end per-query API calls)}
  & 6{,}100\,s (no precomp.) \\
Claude Sonnet 4.6
  & \multicolumn{3}{c}{N/A (end-to-end per-query API calls)}
  & 9{,}415\,s (no precomp.) \\
\bottomrule
\end{tabular}
\caption{Execution time estimates for 1,000 queries with 25 candidates. \textit{Precomputation Time}: GE2: ${\sim}2.0$\,s/query, ${\sim}2.5$\,s/image; Amazon Nova 2: ${\sim}1.8$\,s/query, ${\sim}2.0$\,s/image; the image embedding time is aggregated over 5,000 unique candidate images. \textit{With precomputation}: embeddings are computed once and reused for ranking.}
\label{tab:timing}

\end{table*}

\begin{figure}[!t]
    \centering
    \includegraphics[width=\columnwidth]{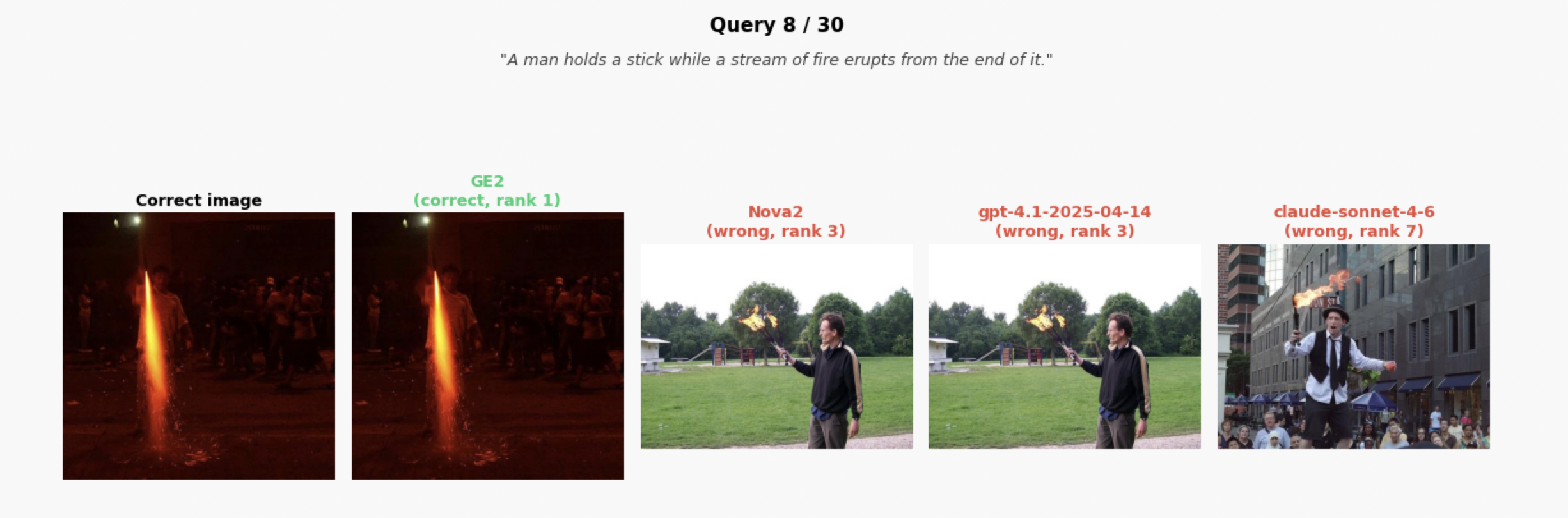}
    \caption{Gemini Embedding 2 model is correct; others incorrect. }
    \label{fig:ge2-result}
\end{figure}

GE2's accuracy parity with LLMs-based rankers suggests that its embedding space is optimized for retrieval, whereas LLMs are generalist models that rank through prompting. The accuracy parity could also be potentially attributed to embedding models' similarity scores computed independently for each query-candidate pair, making them robust to candidate set composition and free from position or distractor biases \citep{wang2023notfair}. However, we minimized this positioning bias in LLMs by randomizing the order per query. In contrast, LLMs have an advantage that they receive all 25 images simultaneously and can reason about relative visual content across candidates before producing a ranking. This cross-candidate reasoning might be an advantage for the LLMs. As a qualitative example, in Figure \ref{fig:ge2-result} we observe that GE2 retrieved the correct image despite it being dark, while Nova 2 and GPT-4.1 selected the same image as their top-ranked image. Additional examples are provided in Appendix \ref{appendix:qualitative observations}.

Amazon Nova 2 trails GE2 by 13.1 percentage points on Recall@1. We attribute this gap to the use of \texttt{GENERIC\_INDEX} as the embedding purpose, which is designed for general vector database creation rather than optimized text-image matching. While this choice reflects realistic vector database deployment scenarios, the Amazon Bedrock API offers a dedicated \texttt{IMAGE\_RETRIEVAL} purpose that may yield stronger performance on text-to-image tasks; we leave this comparison to future work.


\paragraph{Precomputation - the practical case for embedding models.} 
Embedding models are highly suitable for ranking large data with lower latency owing to precomputation, compared to using LLMs as visual rankers which are limited to a smaller candidate set. Without precomputation, embedding model total time is broadly comparable to LLM end-to-end time (depending on candidate pool overlap). Table~\ref{tab:timing} shows that once precomputation is complete, online ranking latency (for 1000 queries and 25 candidates) drops to under 2 seconds for both Gemini Embedding 2 and Amazon Nova 2, compared to 6,100 seconds and 9,415 seconds for GPT-4.1 and Claude Sonnet 4.6, respectively. This speedup is four orders of magnitude faster. For applications requiring sub-second retrieval over a large but fixed set of text and/or images, embedding models are the better choice (especially GE2 because of high recall scores). One critical distinction is that embedding model costs are incurred \textit{once}, when creating the embeddings, and amortized across all subsequent queries. On the other hand, LLM costs scale linearly with every query. LLM-based rankers may be preferable when the image gallery is frequently updated or small, rendering precomputation impractical or its amortized benefit negligible.

\section{Conclusion}
\label{sec:conclusion}
We have presented a comparison of natively
multimodal embedding models (Gemini Embedding 2 and Amazon Nova 2), and LLMs (GPT-4.1 and Claude-Sonnet-4.6) for text-to-image retrieval, using hard-negative candidates from Flickr30k.
Gemini Embedding 2, GPT-4.1, and Claude Sonnet 4.6 achieve statistically indistinguishable retrieval accuracy on all three metrics - Recall@1, Recall@3, and MRR. The practical implication favors multimodal embedding models for low-latency applications. Once embeddings are precomputed, Gemini Embedding 2 and Amazon Nova 2 rank 1,000 queries in under two seconds, compared to 6,100 and 9,415 seconds for GPT-4.1 and Claude Sonnet 4.6, respectively. Extending this comparison to additional datasets and to video and audio modalities are natural directions for future work.

\section*{Limitations}
\label{sec:limitations}

\paragraph{Caption--image correspondence.}
Flickr30k captions describe images at varying levels of specificity, and
a caption from one image may plausibly describe another image in the
dataset, particularly among hard negatives selected for semantic
similarity. This creates ambiguous ground truth for some queries. We
argue this limitation affects all evaluated systems equally, as none is
given privileged information about which image served as the caption's
source.

\paragraph{Hard negatives defined in caption space.}
Candidate sets are hard with respect to caption-level semantics but the "hardness" may change in image embedding space.
A distractor that is semantically close in caption space may or may not
be visually confusable in the embedding or reasoning space of a given model. This is a deliberate design choice, as discussed in Section~\ref{sec:methodology}. Another limitation of selecting hard negatives based on caption similarity is that it may inherently favor embedding-based methods optimized for dense vector representations.

\paragraph{Different inference conditions.}
The two paradigms operate under fundamentally different inference conditions: LLMs receive all 25 candidate images simultaneously and may exploit cross-candidate comparisons before producing a ranking, while embedding models score each image independently without access to other candidates. 

\paragraph{Proprietary model access.}
All four evaluated systems are accessible only through paid APIs, and model weights may be updated by providers between our experiments and any replication attempt.

\paragraph{Candidate set size and generalizability.}
With 25-image candidate sets, our evaluation measures ranking performance rather than full-gallery retrieval. Conclusions about
relative system performance may not generalize to large-scale settings. 

\paragraph{LLM Ranking Prompt and Replicability.}
Changes to LLM prompt for GPT-4.1, and Claude-Sonnet-4.6 may affect the ranking. For reproducibility, we have reported the prompt we created, in Appendix~\ref{appendix:llm_ranking_prompt}.  Additionally, candidates could be presented to LLMs in smaller batches rather than all 25 simultaneously, which may mitigate position bias.

\bibliography{custom}

@techreport{amazon2025novamme,
  title       = {Amazon Nova Multimodal Embeddings: Technical Report and Model Card},
  author      = {{Amazon}},
  year        = {2025},
  month       = {October},
  institution = {Amazon Web Services},
  url         = {https://assets.amazon.science/de/d4/149300334682a464963f01553ffb/nova-mme-technical-report-10.pdf}
}

@inproceedings{radford2021clip,
  title        = {Learning Transferable Visual Models From Natural Language Supervision},
  author       = {Radford, Alec and Kim, Jong Wook and Hallacy, Chris and Ramesh, Aditya
                  and Goh, Gabriel and Agarwal, Sandhini and Sastry, Girish and
                  Askell, Amanda and Mishkin, Pamela and Clark, Jack and others},
  booktitle    = {Proceedings of the 38th International Conference on Machine Learning},
  pages        = {8748--8763},
  year         = {2021},
  organization = {PMLR}
}

@inproceedings{jia2021align,
  title        = {{ALIGN}: Scaling Up Visual and Vision-Language Representation Learning
                  With Noisy Text Supervision},
  author       = {Jia, Chao and Yang, Yinfei and Xia, Ye and Chen, Yi-Ting and Parekh, Zarana
                  and Pham, Hieu and Le, Quoc V. and Sung, Yun-hsuan and Li, Zhen
                  and Duerig, Tom},
  booktitle    = {Proceedings of the 38th International Conference on Machine Learning},
  pages        = {4904--4916},
  year         = {2021},
  organization = {PMLR}
}

@inproceedings{faghri2018vse,
  title     = {{VSE++}: Improving Visual-Semantic Embeddings with Hard Negatives},
  author    = {Faghri, Fartash and Fleet, David J. and Kiros, Jamie Ryan and Fidler, Sanja},
  booktitle = {British Machine Vision Conference (BMVC)},
  year      = {2018}
}

@misc{google2026ge2,
  title  = {{Gemini Embedding 2}: Our First Natively Multimodal Embedding Model},
  author = {{Google DeepMind}},
  year   = {2026},
  note   = {Available at \url{https://blog.google/innovation-and-ai/models-and-research/gemini-models/gemini-embedding-2/}. Accessed March 2026.}
}

@misc{anthropic2024claude,
  title  = {{Claude}: A Family of Large Language Models},
  author = {{Anthropic}},
  year   = {2024},
  note   = {Available at \url{https://www.anthropic.com}}
}

@inproceedings{reimers2019sbert,
  title     = {Sentence-{BERT}: Sentence Embeddings using {Siamese} {BERT}-Networks},
  author    = {Reimers, Nils and Gurevych, Iryna},
  booktitle = {Proceedings of the 2019 Conference on Empirical Methods in
               Natural Language Processing},
  pages     = {3982--3992},
  year      = {2019}
}

@inproceedings{li2023blip2,
  title        = {{BLIP-2}: Bootstrapping Language-Image Pre-training with
                  Frozen Image Encoders and Large Language Models},
  author       = {Li, Junnan and Li, Dongxu and Savarese, Silvio and Hoi, Steven},
  booktitle    = {International Conference on Machine Learning},
  pages        = {19730--19742},
  year         = {2023},
  organization = {PMLR}
}

@inproceedings{dai2023instructblip,
  title     = {{InstructBLIP}: Towards General-Purpose Vision-Language Models
               with Instruction Tuning},
  author    = {Dai, Wenliang and Li, Junnan and Li, Dongxu and Tiong, Anthony Meng Huat
               and Zhao, Junqi and Wang, Weisheng and Li, Boyang and Fung, Pascale
               and Hoi, Steven},
  booktitle = {Advances in Neural Information Processing Systems},
  volume    = {36},
  year      = {2023}
}

@misc{zhu2024mllm,
  author       = {Zhu, Weizhi and others},
  title        = {{MLLM} is a Strong Reranker: Advancing Multimodal 
                  Retrieval-Augmented Generation via Knowledge-Enhanced 
                  Reranking and Noise-Injected Training},
  year         = {2024},
  eprint       = {2407.21439},
  archivePrefix= {arXiv},
  primaryClass = {cs.IR},
  url          = {https://arxiv.org/abs/2407.21439}
}

@inproceedings{sun2023chatgpt,
  title     = {Is {ChatGPT} Good at Search? Investigating Large Language Models as
               Re-Ranking Agents},
  author    = {Sun, Weiwei and Yan, Lingyong and Ma, Xinyu and Wang, Shuaiqiang
               and Ren, Pengjie and Chen, Zhumin and Yin, Dawei and Ren, Zhaochun},
  booktitle = {Proceedings of the 2023 Conference on Empirical Methods in
               Natural Language Processing},
  pages     = {14918--14937},
  year      = {2023}
}

@article{ma2023zero,
  title   = {Zero-Shot Listwise Document Reranking with a Large Language Model},
  author  = {Ma, Xueguang and Zhang, Xinyu and Pradeep, Ronak and Lin, Jimmy},
  journal = {arXiv preprint arXiv:2305.02156},
  year    = {2023}
}

@inproceedings{girdhar2023imagebind,
  title     = {{ImageBind}: One Embedding Space to Bind Them All},
  author    = {Girdhar, Rohit and El-Nouby, Alaaeldin and Liu, Zhuang and Singh, Mannat
               and Alwala, Kalyan Vasudev and Joulin, Armand and Misra, Ishan},
  booktitle = {Proceedings of the IEEE/CVF Conference on Computer Vision and
               Pattern Recognition},
  pages     = {15180--15190},
  year      = {2023}
}

@article{fang2023evaclip,
  title   = {{EVA-CLIP}: Improved Training Techniques for {CLIP} at Scale},
  author  = {Fang, Yuxin and Sun, Quan and Wang, Xinggang and Huang, Tiejun
             and Wang, Xinlong and Cao, Yue},
  journal = {arXiv preprint arXiv:2303.15389},
  year    = {2023}
}

@inproceedings{robinson2021contrastive,
  title     = {Contrastive Learning with Hard Negative Samples},
  author    = {Robinson, Joshua and Chuang, Ching-Yao and Sra, Suvrit and Jegelka, Stefanie},
  booktitle = {International Conference on Learning Representations},
  year      = {2021}
}

@article{mcnemar1947note,
  title   = {Note on the Sampling Error of the Difference Between Correlated
             Proportions or Percentages},
  author  = {McNemar, Quinn},
  journal = {Psychometrika},
  volume  = {12},
  number  = {2},
  pages   = {153--157},
  year    = {1947}
}

@book{efron1993bootstrap,
  title     = {An Introduction to the Bootstrap},
  author    = {Efron, Bradley and Tibshirani, Robert J.},
  year      = {1993},
  publisher = {Chapman and Hall/CRC}
}

@inproceedings{lin2014coco,
  title     = {Microsoft {COCO}: Common Objects in Context},
  author    = {Lin, Tsung-Yi and Maire, Michael and Belongie, Serge and
               Hays, James and Perona, Pietro and Ramanan, Deva and
               Doll{\'a}r, Piotr and Zitnick, C. Lawrence},
  booktitle = {European Conference on Computer Vision (ECCV)},
  pages     = {740--755},
  year      = {2014},
  publisher = {Springer}
}

@article{wilcoxon1945individual,
  title   = {Individual Comparisons by Ranking Methods},
  author  = {Wilcoxon, Frank},
  journal = {Biometrics Bulletin},
  volume  = {1},
  number  = {6},
  pages   = {80--83},
  year    = {1945}
}

@article{young2014flickr30k,
  title   = {From image descriptions to visual denotations: New similarity
             metrics for semantic inference over event descriptions},
  author  = {Young, Peter and Lai, Alice and Hodosh, Micah and Hockenmaier, Julia},
  journal = {Transactions of the Association for Computational Linguistics},
  volume  = {2},
  pages   = {67--78},
  year    = {2014},
  publisher = {MIT Press},
  doi     = {10.1162/tacl_a_00166}
}

@article{wang2023notfair,
  title   = {Large Language Models are not Fair Evaluators},
  author  = {Wang, Peiyi and Li, Lei and Chen, Liang and Cai, Zefan and
             Zhu, Dawei and Lin, Binghuai and Cao, Yunbo and Liu, Qi and
             Liu, Tianyu and Sui, Zhifang},
  journal = {arXiv preprint arXiv:2305.17926},
  year    = {2023}
}

@inproceedings{jiang2024vlm2vec,
  title     = {{VLM2Vec}: Training Vision-Language Models for Massive Multimodal Embedding Tasks},
  author    = {Jiang, Ziyan and Meng, Rui and Yang, Xinyi and Castillo, Semih Yavuz and Yavuz, Semih and Chen, Wenhu},
  booktitle = {International Conference on Learning Representations},
  year      = {2025},
  url       = {https://arxiv.org/abs/2410.05160}
}

@misc{jiang2024mmembed,
  author    = {Jiang, Hongwei and Lian, Tengyu and others},
  title     = {{MM-Embed}: Universal Multimodal Retrieval with Multimodal {LLM}s},
  year      = {2024},
  eprint    = {2411.02571},
  archivePrefix = {arXiv},
  primaryClass  = {cs.IR},
  url       = {https://arxiv.org/abs/2411.02571}
}

@techreport{openai2025gpt41,
  author      = {{OpenAI}},
  title       = {{GPT-4.1} System Card},
  institution = {OpenAI},
  year        = {2025},
  url         = {https://openai.com/index/gpt-4-1/}
}
\appendix

\section{Details of Evaluated Systems}
\label{appendix:systems_configuration}
 
\paragraph{System 1: Gemini Embedding 2 (Gemini Embedding 2).} We embed each query caption using the \texttt{gemini-embedding-2-preview}
model with \texttt{task\_type=RETRIEVAL\_QUERY} and embed each candidate
image with \texttt{task\_type=RETRIEVAL\_DOCUMENT}, using
3{,}072-dimensional output. We compute cosine similarity between the caption embedding and each of the 25 candidate image embeddings, and rank the
candidates from highest to lowest similarity.

\paragraph{System 2: Amazon Nova 2 - Multimodal Embedding 2.} We embed each query caption using the \texttt{amazon.nova-2-multimodal-embeddings-v1:0}
model with \texttt{embedding\_purpose=GENERIC\_INDEX} and also embed each candidate
image with \texttt{embedding\_purpose=GENERIC\_INDEX}, using 3,072-dimensional output. We compute cosine similarity between the caption embedding and each of the 25 candidate image embeddings and rank the
candidates from highest to lowest similarity.
 
\paragraph{System 3: GPT-4.1}
We send the query caption and all 25 candidate images in a single prompt (zero-shot)
to \texttt{gpt-4.1-2025-04-14} (latest snapshot at time of experiment). Images are presented in randomized order per query to control for position bias
\citep{wang2023notfair}, numbered 1 through 25. The prompt instructs the model to return a ranking of all 25 images by their relevance to the caption, formatted as a list of image numbers. Randomization is applied
independently per query, and the same random seed is used across both LLM conditions for reproducibility.
 
\paragraph{System 4: Claude Sonnet 4.6}
We replicate the same procedure as GPT-4.1 for the
\texttt{claude-sonnet-4-6 model} \cite{anthropic2024claude}.

\section{Metrics with Confidence Intervals}
\label{appendix:metrics_with_ci}


Table~\ref{tab:full_results} reports the complete metric results for all
four systems with 95\% bootstrap confidence intervals (1,000 resamples)
and total ranking execution time for 1,000 queries.

\begin{table*}[h]
\centering
\small
\begin{tabular}{lcccccc}
\toprule
\textbf{System} & \textbf{R@1} & \textbf{95\% CI} & \textbf{R@3}
                & \textbf{95\% CI} & \textbf{MRR} & \textbf{95\% CI} \\
\midrule
GE2
  & 0.804 & [0.780, 0.829]
  & 0.921 & [0.905, 0.937]
  & 0.871 & [0.854, 0.888] \\
Nova2
  & 0.673 & [0.644, 0.703]
  & 0.844 & [0.821, 0.867]
  & 0.774 & [0.753, 0.796] \\
GPT-4.1
  & 0.789 & [0.763, 0.816]
  & 0.915 & [0.898, 0.932]
  & 0.859 & [0.841, 0.877] \\
Claude Sonnet 4.6
  & 0.796 & [0.772, 0.823]
  & 0.903 & [0.886, 0.921]
  & 0.857 & [0.839, 0.876] \\
\bottomrule
\end{tabular}
\caption{Full results for all systems on 1,000 queries with 25 hard-negative
candidates per query. Bootstrap CIs are 95\% intervals computed over
1,000 resamples. Random baseline values are theoretical (1/25 for R@1,
3/25 for R@3, harmonic mean for MRR).}
\label{tab:full_results}
\end{table*}


\section{LLM Ranking Prompt}
\label{appendix:llm_ranking_prompt}
Here's the LLM Ranking prompt used:

\texttt{ "system": "You are a precise image retrieval system. Given a text query and a set of candidate images, rank ALL images from most to least relevant. Relevance means how well the image matches the specific subjects, actions, setting, and details described in the query. You must return a JSON array containing every image ID exactly once. Output ONLY the JSON array — no explanation, no commentary, nothing else."
}

\texttt{
"intro\_template": "Text query: {query\_caption}. You are given {n\_candidates} candidate images, each labeled with an image ID shown before it. For each image, consider: (1) Does it show the same subjects or people described in the query? (2) Does it show the same action or activity? (3) Does it match the setting, location, or scene? (4) Does it contain the specific objects or details mentioned? Images:"
}

\texttt{
"outro": "Now return a JSON array of ALL {n\_candidates} image IDs ranked from most relevant (index 0) to least relevant (last index). Every ID must appear exactly once. If two images seem equally relevant, use your best judgment to order them — do not omit any ID. Provide JSON array only:",
}

\section{More Hard Negative Dataset Examples}
\label{appendix:more_hard_negatives}

\begin{figure}[h]
\centering
\begin{minipage}{\linewidth}
  \includegraphics[width=\linewidth]{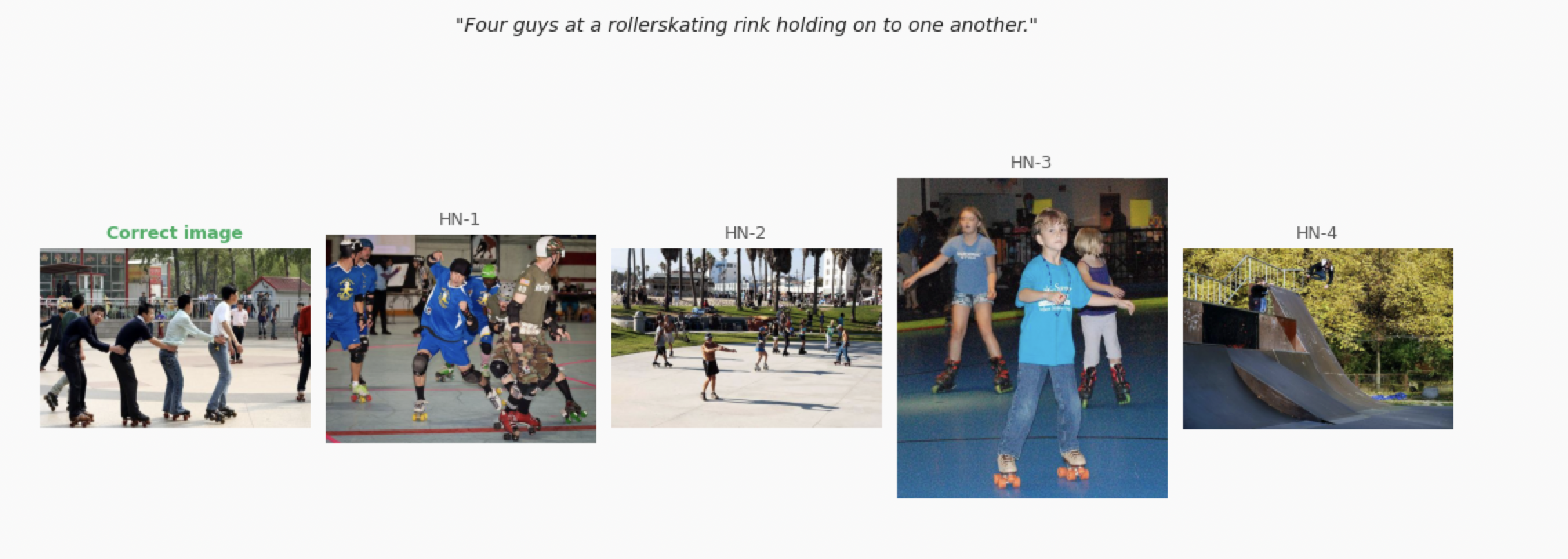}
\end{minipage}
\caption{Hard negative example 1. The query caption describes a specific
scene; candidate images share overlapping visual and semantic elements
that make the retrieval task non-trivial. }
\label{fig:hn_example1}
\end{figure}

\begin{figure}[h]
\centering
\begin{minipage}{\linewidth}
  \includegraphics[width=\linewidth]{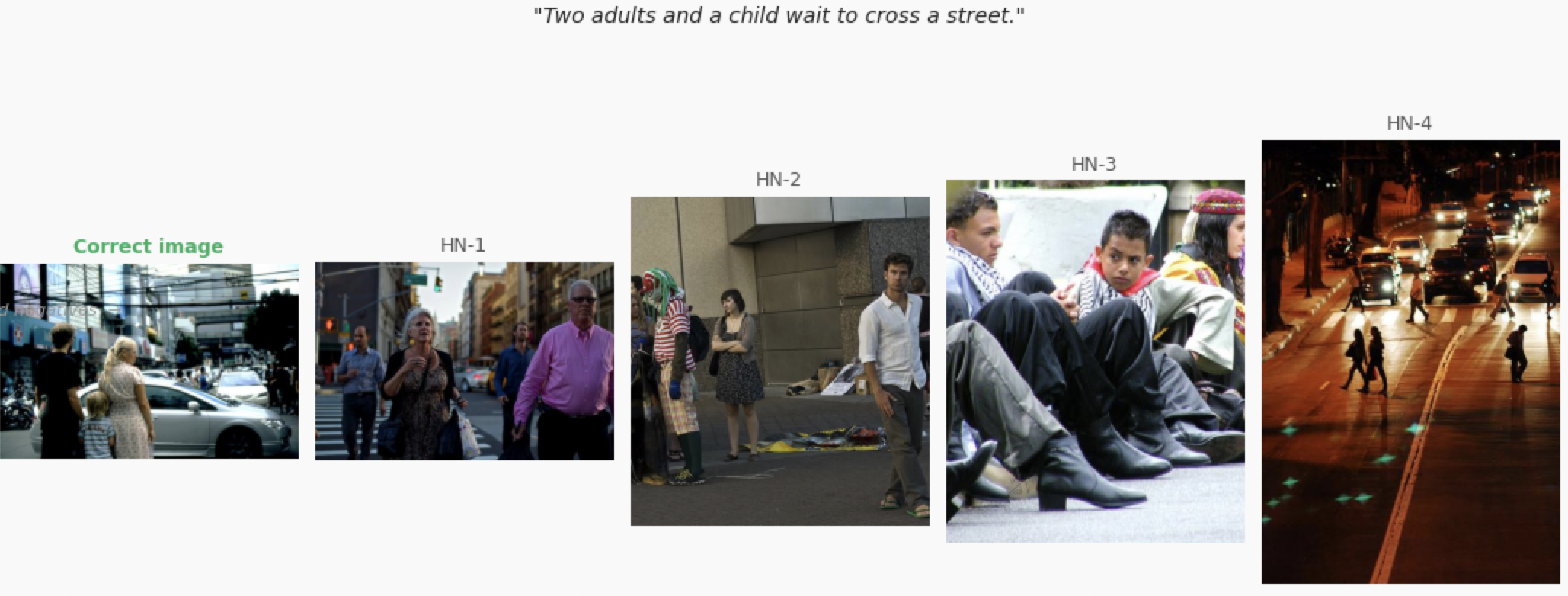}
\end{minipage}
\caption{Hard negative example 2.}
\label{fig:hn_example2}
\end{figure}

\begin{figure}[h]
\centering
\begin{minipage}{\linewidth}
  \includegraphics[width=\linewidth]{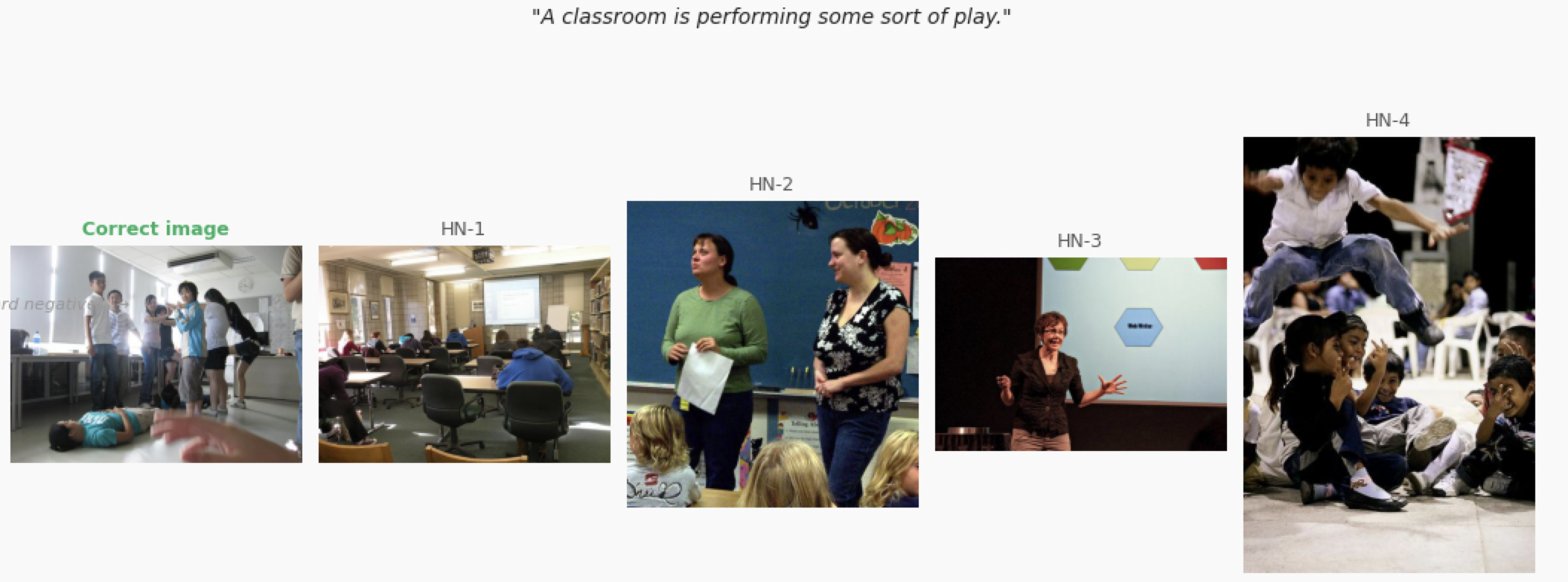}
\end{minipage}
\caption{Hard negative example 3.}
\label{fig:hn_example3}
\end{figure}

Figures~\ref{fig:hn_example1}--\ref{fig:hn_example3} illustrate representative
hard negative candidate sets. Each figure shows the query caption, the
correct image (green border), and the top-$k$ most semantically similar
images as selected by \texttt{all-mpnet-base-v2} caption similarity. These
examples demonstrate that the hard negatives are visually and semantically
confusable with the correct image, validating the difficulty of the
evaluation task.

\section{Contingency Tables}
\label{appendix:contingency}
Table~\ref{tab:contingency} shows the R@1 contingency table for GE2 versus GPT-4.1. Of the 1,000 queries, 716 are answered correctly by both systems, while 88 are answered correctly by GE2 alone and 73 by GPT-4.1 alone. 

\begin{table}[H]
\centering
\small
\begin{tabular}{lcc}
\toprule
 & \textbf{GPT-4.1 correct} & \textbf{GPT-4.1 wrong} \\
\midrule
\textbf{GE2 correct} & 716 & 88 \\
\textbf{GE2 wrong}   & 73  & 123 \\
\bottomrule
\end{tabular}
\caption{R@1 contingency table for Gemini Embedding 2 vs.\ GPT-4.1 (1,000 queries). Off-diagonal cells represent qualitatively distinct failure modes analysed in Section~\ref{sec:results-and-discussion}. Contingency tables for all other system pairs are in Appendix~\ref{appendix:contingency}.}
\label{tab:contingency}
\end{table}

Tables~\ref{tab:cont_ge2_claude}--\ref{tab:cont_claude_nova} report R@1
contingency tables for the remaining system pairs not shown in the main
text. Off-diagonal cells indicate queries where the two systems disagree,
forming the basis for qualitative error analysis.

\begin{table}[H]
\centering
\small
\begin{tabular}{lcc}
\toprule
 & \textbf{Claude correct} & \textbf{Claude wrong} \\
\midrule
\textbf{GE2 correct} & 729 & 75 \\
\textbf{GE2 wrong}   & 67  & 129  \\
\bottomrule
\end{tabular}
\caption{R@1 contingency table: GE2 vs.\ Claude Sonnet 4.6 (1,000 queries).}
\label{tab:cont_ge2_claude}
\end{table}

\begin{table}[H]
\centering
\small
\begin{tabular}{lcc}
\toprule
 & \textbf{Claude correct} & \textbf{Claude wrong} \\
\midrule
\textbf{GPT-4.1 correct} & 732 & 57  \\
\textbf{GPT-4.1 wrong}   & 64  & 147 \\
\bottomrule
\end{tabular}
\caption{R@1 contingency table: GPT-4.1 vs.\ Claude Sonnet 4.6 (1,000 queries).}
\label{tab:cont_gpt_claude}
\end{table}

\begin{table}[H]
\centering
\small
\begin{tabular}{lcc}
\toprule
 & \textbf{Nova2 correct} & \textbf{Nova2 wrong} \\
\midrule
\textbf{GE2 correct} & 634 & 170 \\
\textbf{GE2 wrong}   & 39  & 157 \\
\bottomrule
\end{tabular}
\caption{R@1 contingency table: GE2 vs.\ Nova2 (1,000 queries).}
\label{tab:cont_ge2_nova}
\end{table}

\begin{table}[H]
\centering
\small
\begin{tabular}{lcc}
\toprule
 & \textbf{GPT-4.1 correct} & \textbf{GPT-4.1 wrong} \\
\midrule
\textbf{Nova2 correct} & 617 & 56 \\
\textbf{Nova2 wrong}   & 172  & 155 \\
\bottomrule
\end{tabular}
\caption{R@1 contingency table: GPT-4.1 vs.\ Nova2 (1,000 queries).}
\label{tab:cont_gpt_nova}
\end{table}

\begin{table}[H]
\centering
\small
\begin{tabular}{lcc}
\toprule
 & \textbf{Claude correct} & \textbf{Claude wrong} \\
\midrule
\textbf{Nova2 correct} & 624 & 49 \\
\textbf{Nova2 wrong}   & 172  & 155 \\
\bottomrule
\end{tabular}
\caption{R@1 contingency table: Claude Sonnet 4.6 vs.\ Nova2 (1,000 queries).}
\label{tab:cont_claude_nova}
\end{table}

\section{Qualitative Observation}
\label{appendix:qualitative observations}
\begin{figure*}[!t]
    \centering
    \includegraphics[width=0.8\textwidth]{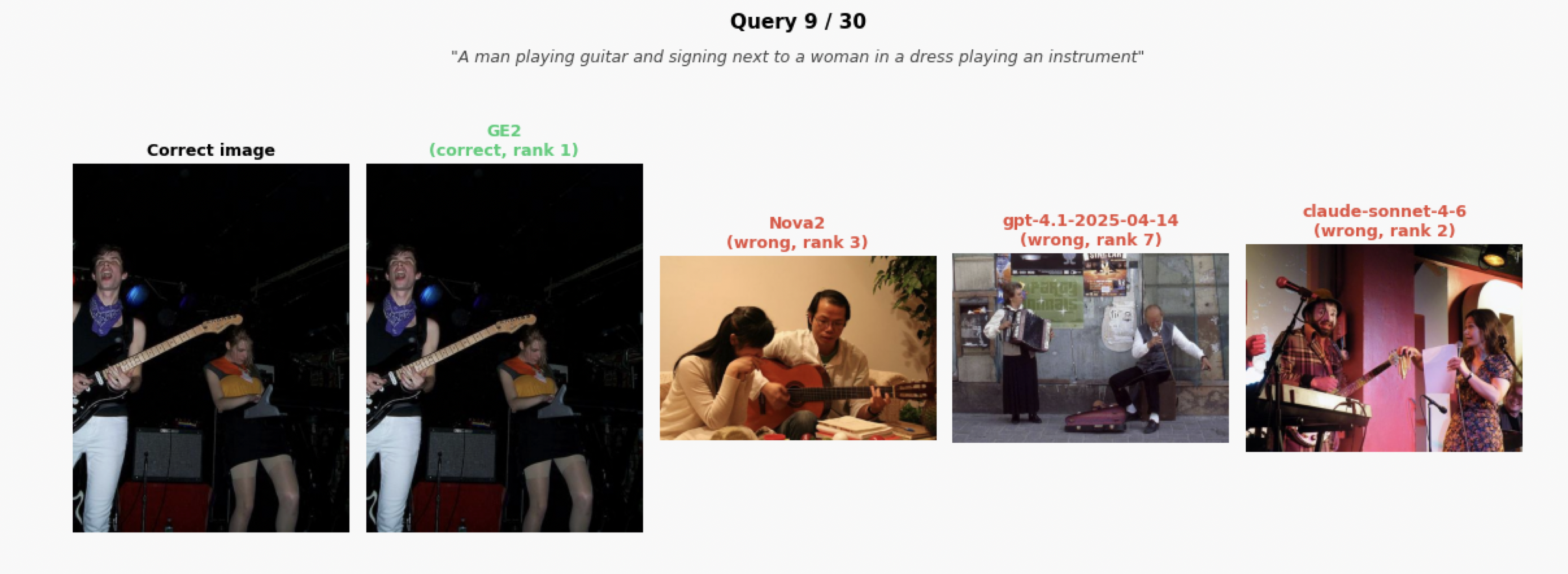}
    \caption{Another example where Gemini Embedding 2 model is correct; others incorrect. }
    \label{fig:ge2_close_with_claude}
\end{figure*}

\begin{figure*}[!t]
    \centering
    \includegraphics[width=0.8\textwidth]{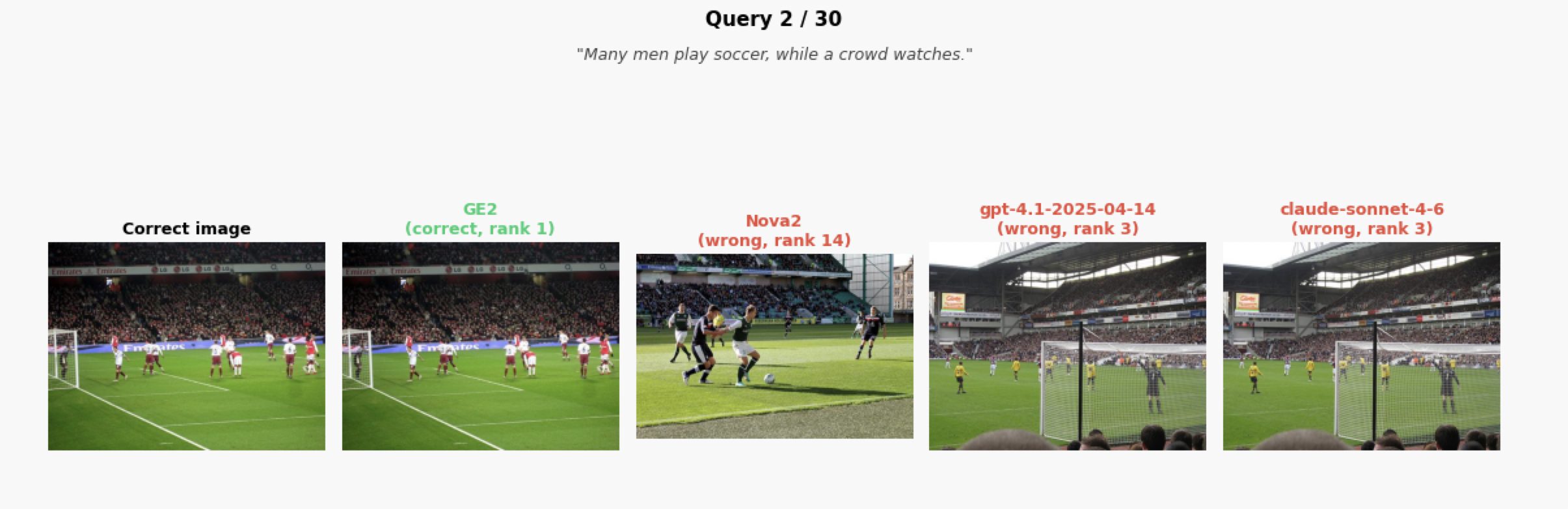}
    \caption{An example where all models' retrieval reasonably close to the query. }
    \label{fig:ge3-result}
\end{figure*}

Each figure shows the query caption, the correct image, and the top-ranked
image returned by each system. Green text indicate a correct rank-1
prediction; red text indicate an incorrect rank-1 prediction.
Figure \ref{fig:ge2_close_with_claude} is another example where GE2 selected the correct image despite it being dark. Figure \ref{fig:ge3-result} shows how close the systems are to each other in terms of retrieval for certain queries. Even though GE2 was technically correct, the other systems also picked "reasonably correct" images (based on human vision). This illustrates that the differences in retrieval capability of each system may be lesser than what the quantitative metrics convey. 

\begin{figure*}[!t]
    \centering
    \includegraphics[width=0.8\textwidth]{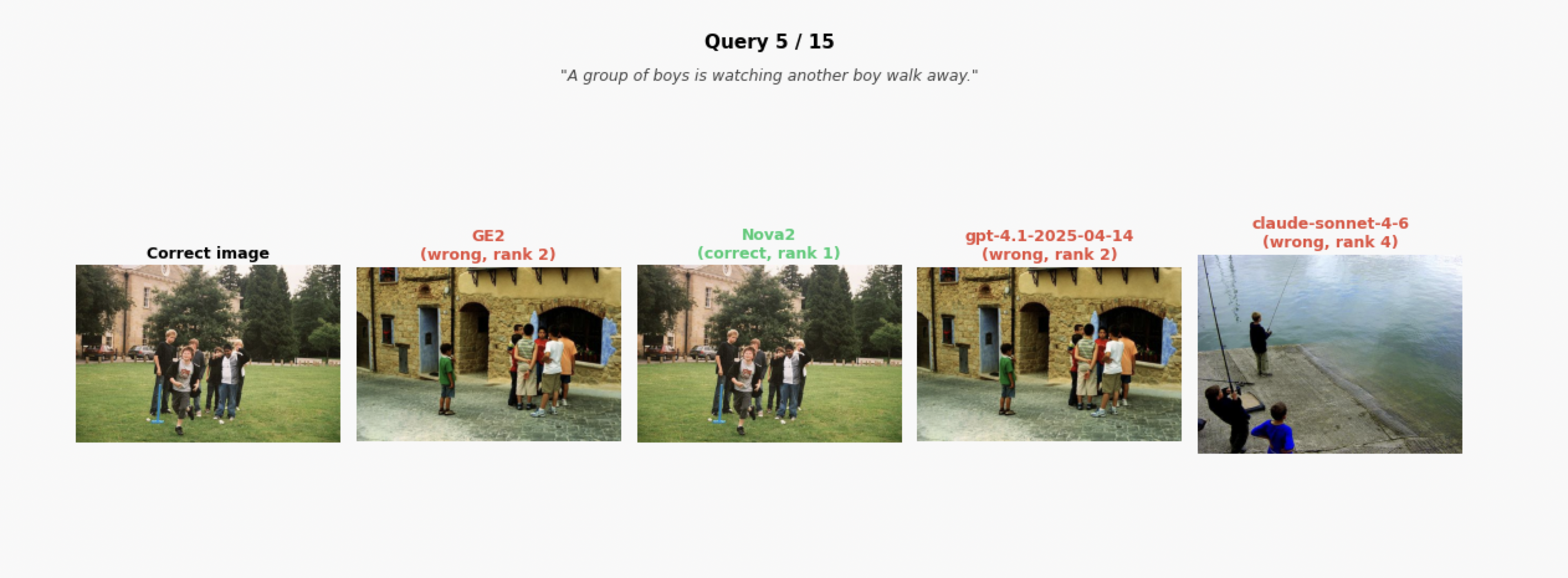}
    \caption{Amazon Nova 2 results are correct; Other models' results are incorrect. }
    \label{fig:nova-result}
\end{figure*}

\begin{figure*}[!t]
    \centering
    \includegraphics[width=0.8\textwidth]{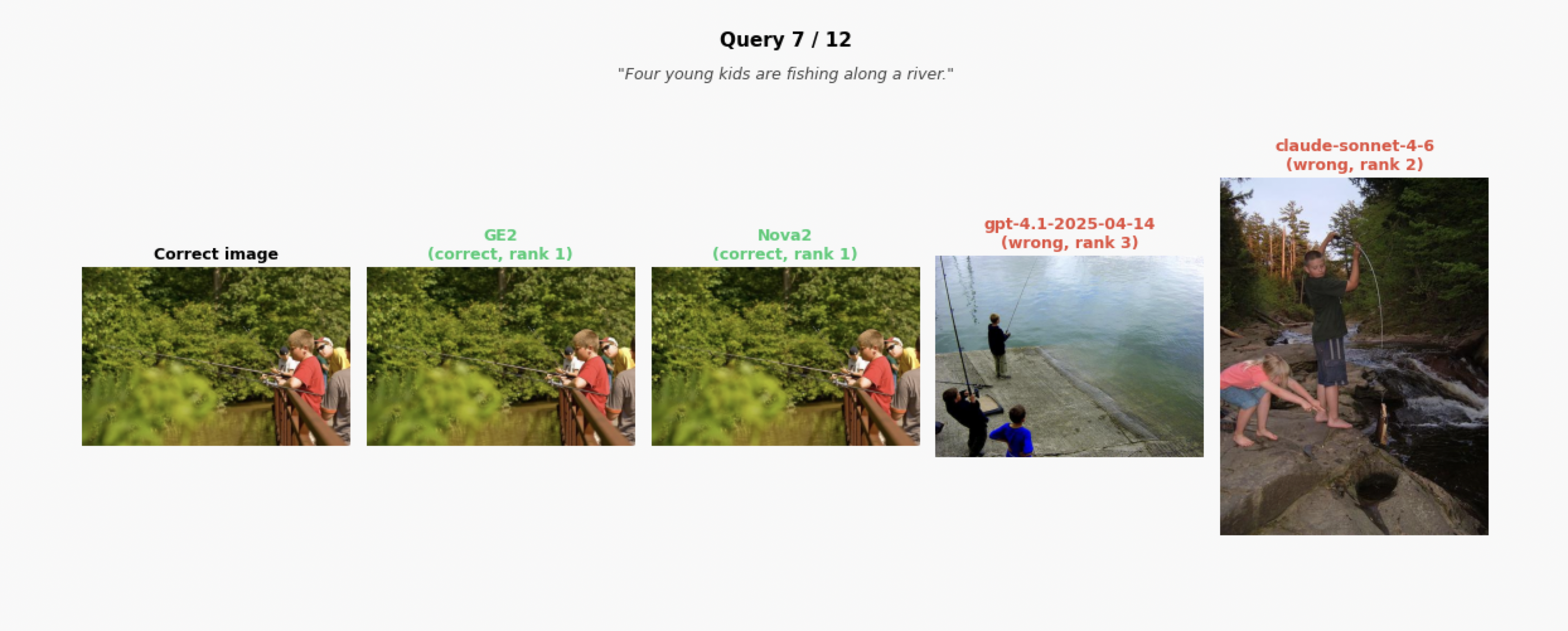}
    \caption{Multimodal embedding models are correct; LLM Rankers incorrect. }
    \label{fig:mm-result}
\end{figure*}

\begin{figure*}[!t]
    \centering
    \includegraphics[width=0.8\textwidth]{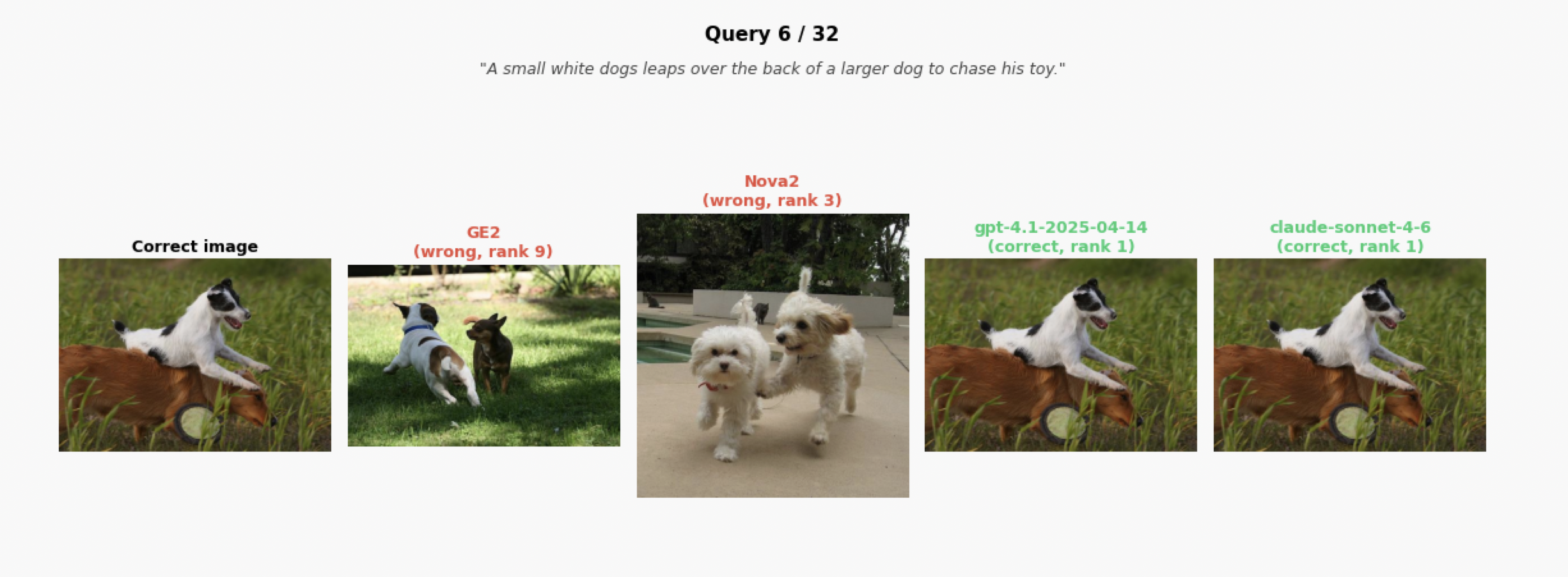}
    \caption{LLM models' results are correct; Embedding models' results are incorrect. }
    \label{fig:llm-result}
\end{figure*}

Interestingly, Figure \ref{fig:nova-result} shows an example when Nova 2 selected the correct image while all others were incorrect. This indicates that Nova 2 may be better for certain queries, which can be explored further. Figure \ref{fig:mm-result} and \ref{fig:llm-result} show where only the Multimodal Embedding models were correct and only LLMs were correct, respectively. 
The following figures provide more examples on the performance of evaluated systems.

\subsection{GE2 correct, all LLMs wrong}
\label{appendix:ge2_only}

\begin{figure*}[h]
\centering
\begin{minipage}[t]{0.48\linewidth}
  \includegraphics[width=\linewidth]{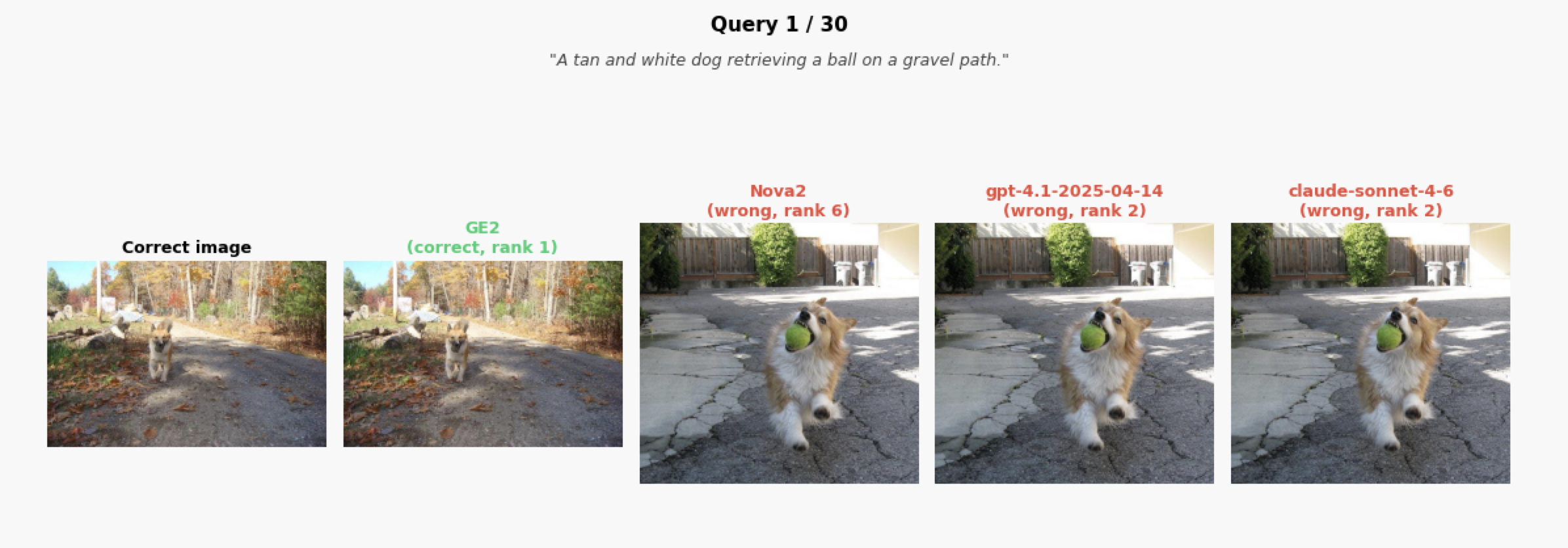}
  \subcaption{Example 1}
\end{minipage}
\hfill
\begin{minipage}[t]{0.48\linewidth}
  \includegraphics[width=\linewidth]{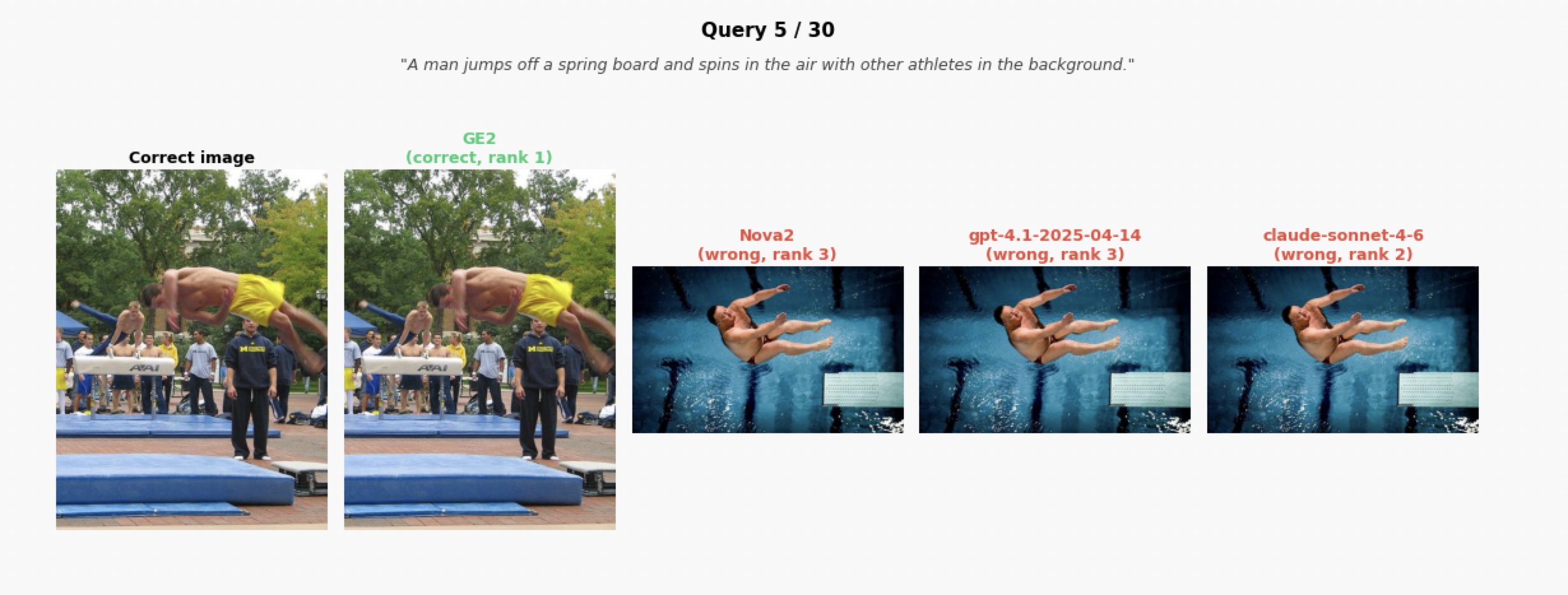}
  \subcaption{Example 2}
\end{minipage}
\caption{Queries where GE2 is correct and both LLMs (GPT-4.1 and Claude
Sonnet) are wrong on R@1. Green = correct rank-1, red = incorrect rank-1.}
\label{fig:ge2_only}
\end{figure*}

Figure~\ref{fig:ge2_only} shows queries where GE2 ranked the correct image
first while GPT-4.1 and Claude Sonnet 4.6 both failed. These cases tend to
involve fine-grained visual similarity where the correct image shares
compositional structure or low-level visual properties with the distractors
that GE2's embedding captures but LLMs miss when reasoning jointly over
25 candidates.

\subsection{Nova2 correct, all other systems wrong}
\label{appendix:nova2_only}

\begin{figure*}[h]
\centering
\begin{minipage}[t]{0.48\linewidth}
  \includegraphics[width=\linewidth]{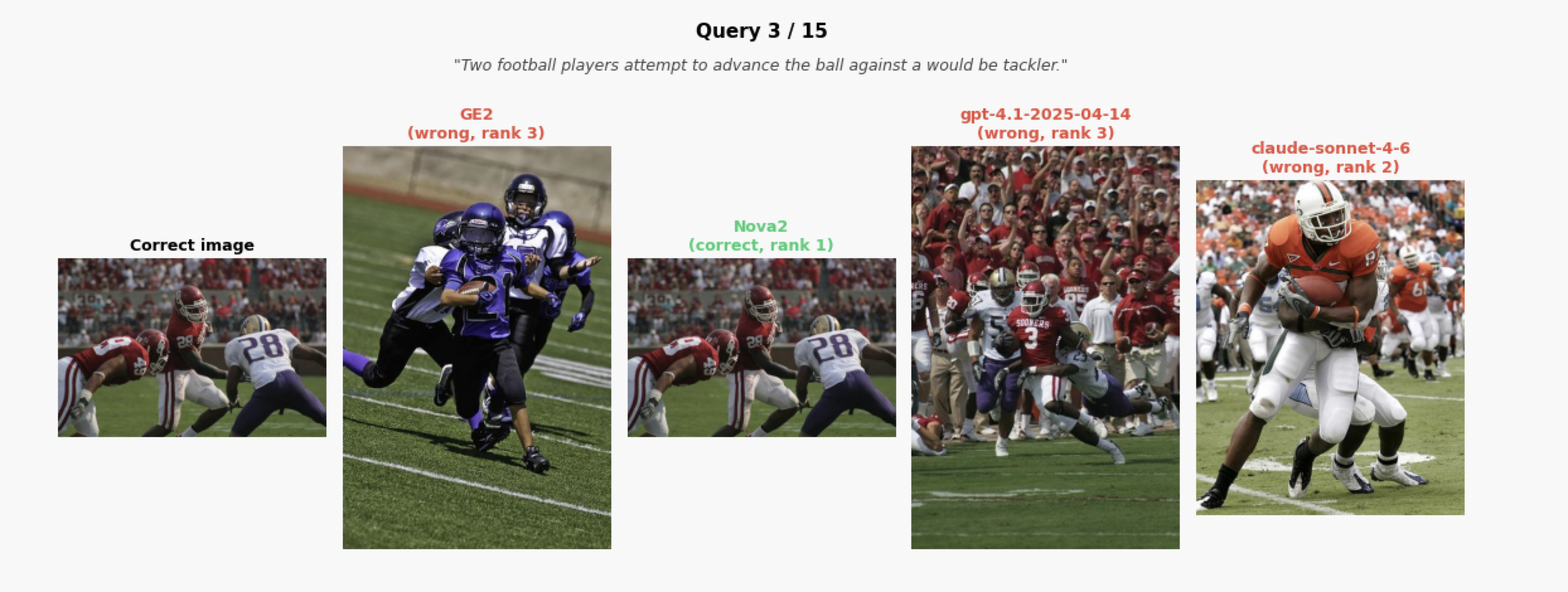}
  \subcaption{Example 1}
\end{minipage}
\hfill
\begin{minipage}[t]{0.48\linewidth}
  \includegraphics[width=\linewidth]{Nova2_correct_only.png}
  \subcaption{Example 2}
\end{minipage}
\caption{Queries where Nova2 is correct and all other systems are wrong on
R@1.}
\label{fig:nova2_only}
\end{figure*}

Figure~\ref{fig:nova2_only} shows the rare queries where Nova2 ranked
correctly while GE2, GPT-4.1, and Claude Sonnet 4.6 all failed. Given Nova2's
overall lower performance, these cases are uncommon but analytically
interesting as they may reflect query types where GENERIC\_INDEX embeddings
capture complementary signal.

\subsection{Both LLMs correct, both embedding models wrong}
\label{appendix:llm_only}

\begin{figure*}[h]
\centering
\begin{minipage}[t]{0.48\linewidth}
  \includegraphics[width=\linewidth]{LLMs_correct_only.png}
  \subcaption{Example 1}
\end{minipage}
\hfill
\begin{minipage}[t]{0.48\linewidth}
  \includegraphics[width=\linewidth]{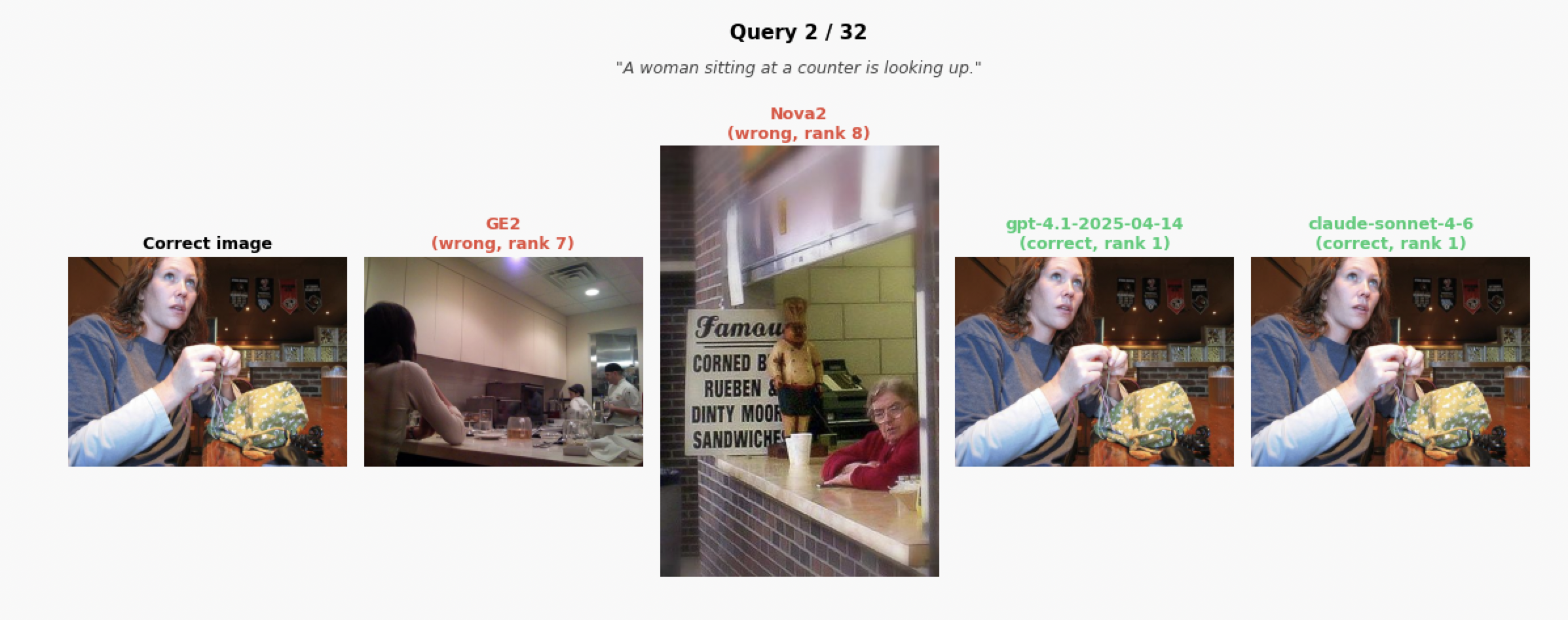}
  \subcaption{Example 2}
\end{minipage}
\caption{Queries where both LLMs (GPT-4.1 and Claude Sonnet 4.6) are correct
and both embedding models (GE2 and Nova2) are wrong on R@1.}
\label{fig:llm_only}
\end{figure*}

Figure~\ref{fig:llm_only} shows queries where both GPT-4.1 and Claude
Sonnet ranked correctly while both GE2 and Nova2 failed. These cases
illustrate the advantage of cross-candidate joint reasoning: the LLM can
compare all 25 images simultaneously and apply semantic or relational
understanding that independent embedding scoring cannot replicate.

\subsection{Both embedding models correct, both LLMs wrong}
\label{appendix:embedding_only}

\begin{figure*}[p]
\vspace*{-0.3cm}
\centering
\begin{minipage}[t]{0.48\linewidth}
  \includegraphics[width=\linewidth]{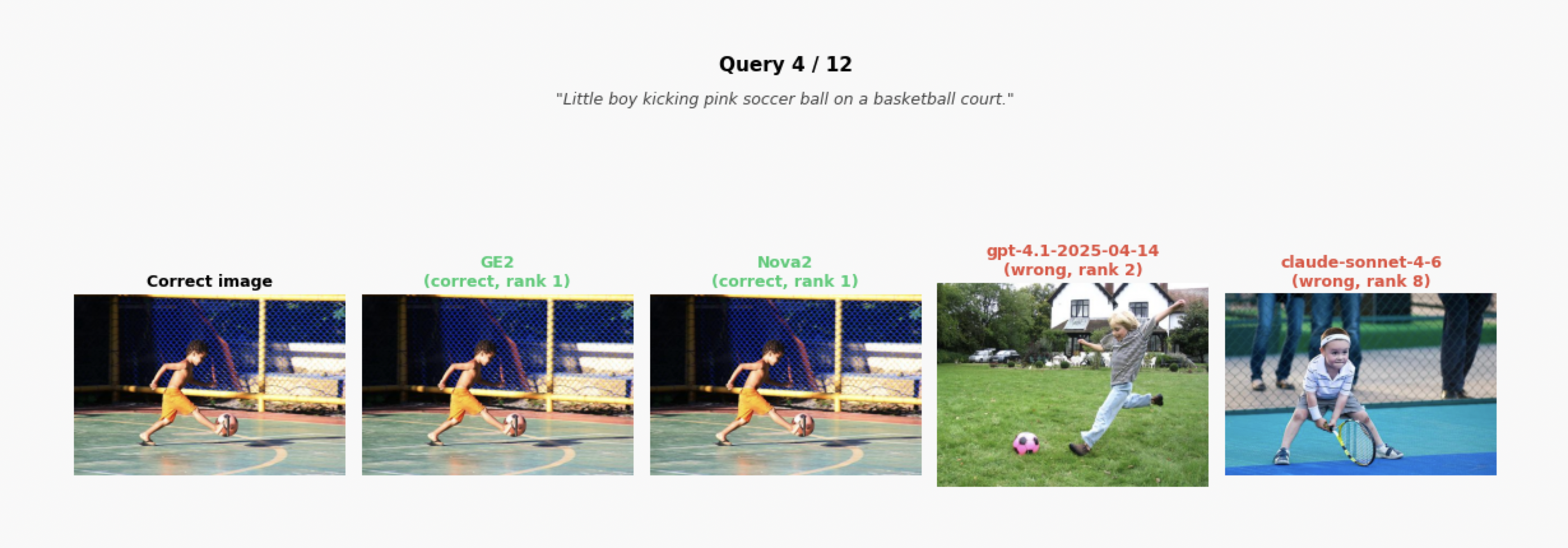}
  \subcaption{Example 1}
\end{minipage}
\hfill
\begin{minipage}[t]{0.48\linewidth}
  \includegraphics[width=\linewidth]{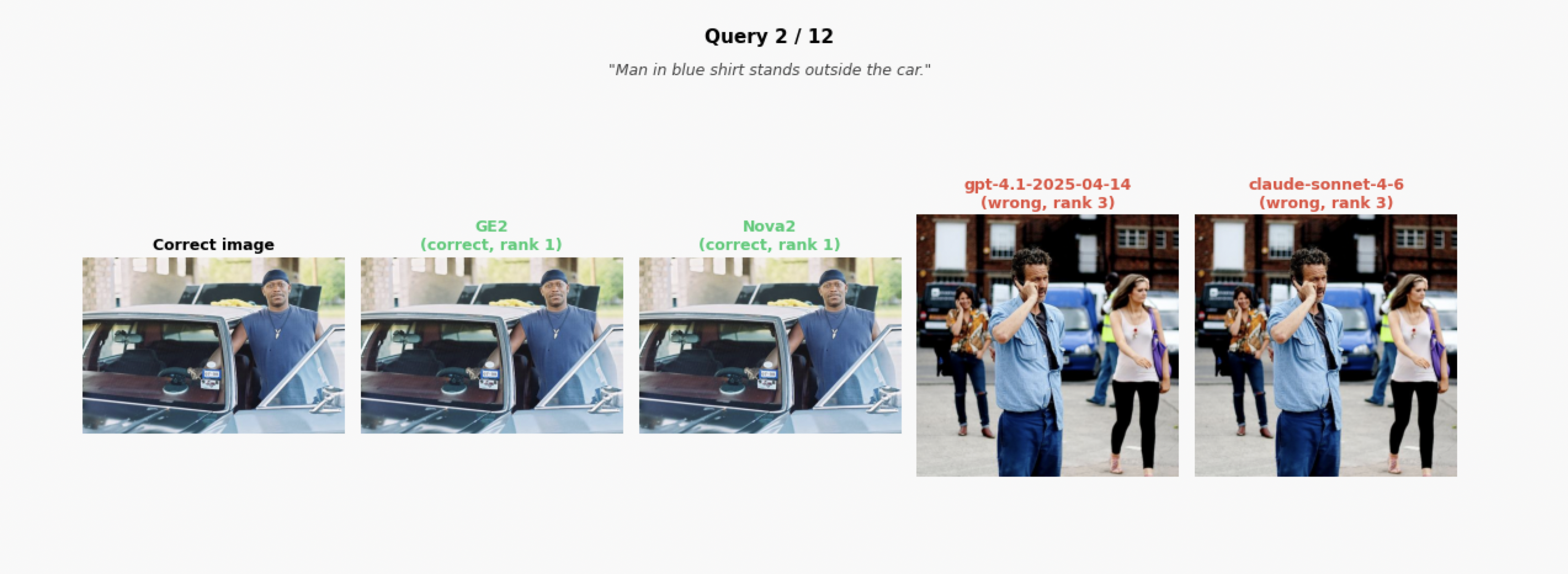}
  \subcaption{Example 2}
\end{minipage}
\caption{Queries where both embedding models (GE2 and Nova2) are correct
and both LLMs (GPT-4.1 and Claude Sonnet 4.6) are wrong on R@1.}
\label{fig:embedding_only}
\end{figure*}

Figure~\ref{fig:embedding_only} shows queries where both GE2 and Nova2
ranked correctly while both LLMs failed. These cases are particularly
informative as they suggest that dense cross-modal embedding captures
visual correspondence that joint visual reasoning consistently misses,
even when the pattern holds across model families.

\end{document}